\documentclass[letterpaper]{article} 
\usepackage{aaai24}
\nocopyright 
\usepackage{times}  
\usepackage{helvet}  
\usepackage{courier}  
\usepackage[hyphens]{url}  
\usepackage{graphicx} 
\usepackage{natbib}  
\usepackage{caption} 
\usepackage{algorithm}
\usepackage{algorithmic}

\usepackage{newfloat}
\usepackage{listings}
\DeclareCaptionStyle{ruled}{labelfont=normalfont,labelsep=colon,strut=off} 
\floatstyle{ruled}
\newfloat{listing}{tb}{lst}{}
\floatname{listing}{Listing}
\title{TMPNN: High-Order Polynomial Regression Based on Taylor Map Factorization}
\author {
    Andrei Ivanov\textsuperscript{\rm 1},
    Stefan Maria Ailuro\textsuperscript{\rm 2},
}
\affiliations {
    \textsuperscript{\rm 1}05x.andrey@gmail.com, 
    \textsuperscript{\rm 2}ailuro.sm@gmail.com
}

\usepackage{bibentry}

\usepackage{amsmath}
\usepackage{amssymb}
\newcommand{\XX}{\mathbf{X}}

\newcommand{\YY}{\mathbf{Y}}
\newcommand{\ZZ}{\mathbf{Z}}

\newcommand{\MM}{\mathcal{M}}
\newcommand{\x}{\mathbf{x}}
\newcommand{\y}{\mathbf{y}}

\usepackage[table]{xcolor}
\newcommand\colb{\cellcolor{blue!10}}
\newcommand\colr{\cellcolor{red!20}}

\definecolor{gold}{HTML}{E1D701}
\definecolor{silver}{HTML}{C0C0C0}
\definecolor{bronze}{HTML}{CD7F32}

\usepackage{multirow}
\begin{document}

\maketitle

\begin{abstract}
Polynomial regression is widely used and can help to express nonlinear patterns. However, considering very high polynomial orders may lead to overfitting and poor extrapolation ability for unseen data. The paper presents a method for constructing a high-order polynomial regression based on the Taylor map factorization. This method naturally implements multi-target regression and can capture internal relationships between targets. Additionally, we introduce an approach for model interpretation in the form of systems of differential equations. By benchmarking on UCI open access datasets, Feynman symbolic regression datasets, and Friedman-1 datasets, we demonstrate that the proposed method performs comparable to the state-of-the-art regression methods and outperforms them on specific tasks.
\end{abstract}

\section{Introduction and Related Works}

We consider the regression problem as finding the mapping $f$ for feature vector $\XX = \{\x_1, \x_2, \ldots, \x_n\}\in \mathcal R^{n}$ of $n$ input real-valued variables to the dependent target vector $\YY = \{\y_1, \y_2, \ldots, \y_m\}\in \mathcal R^{m}$ of $m$ output real-valued variables:
\begin{equation}
    \label{reg}
    f:\;\{\x_1, \x_2, \ldots, \x_n\}\;\rightarrow\;\{\y_1, \y_2, \ldots, \y_m\}.
\end{equation}

The are two groups of methods to solve this problem.The first one is regular polynomial regression
\begin{equation}
\label{polreg}
    \y_l = w_0 + \sum_{i}w_i\x_i + \sum_{i,j}w_{ij}\x_i\x_j + \ldots\\
\end{equation}
Commonly, linear terms are not enough for complex datasets, while more nonlinear interactions are introduced in \eqref{polreg}, the higher the risk of overfitting. To fit this model efficiently, most authors suggest either optimal strategies for the selection of the nonlinear terms \cite{Stepwise, Hofwing, dette1995, Fan, Hierarchical, Pakdemirli} or different factorization schemes \cite{Freudenthaler, FM}.



The second group of methods is machine learning models like Support Vector Regression (SVR), Gaussian Process Regression (GPR), Random Forest Regression (RFR), or neural networks (NN). These black-box models were initially developed for interpolation problems. GPR with the Gaussian kernel regresses to the mean function when extrapolating far enough from training. Tree-based methods presume the output will be constant values if predictions are made outside the range of the original set. Various techniques, such as kernel-based modified GPR \cite{wilson13} or regression-enhanced RFR \cite{RERF}, are introduced to improve the extrapolation property of these models.

The state-of-the-art models for regression problem \eqref{reg} are CatBoost as a boosting method on decision trees \cite{CatBoost}, TabNet for deep NN \cite{TabNet}, and DNNR for Differential Nearest Neighbors Regression \cite{DNNR}.

The paper introduces a new regression model based on the Taylor map factorization that can be implemented as a polynomial neural network (PNN) with shared weights. The earliest and most relevant to our research is the paper of authors \cite{neco}, where the connection between the system of ordinary differential equations (ODEs) and polynomial neural network (PNN) is introduced. Further, the PNN architectures were also widely highlighted in the literature \cite{Oh}. \cite{ref10} proposes a polynomial neural architecture approximating differential equations. The Legendre polynomial is chosen as a basis by \cite{ref11}. \cite{TMPNN} suggest an algorithm for translating ODEs to PNN without training. \cite{wu2022extrapolation} examined the extrapolation ability for PNNs in both simple regression tasks and more complicated computer vision problems. \cite{Fronk_2023} applied PNN for learning ODEs.


Following these works, we propose an approach for the general-purpose multi-target regression task. In contrast to traditional techniques, where several single-output models are either used independently or chained together \cite{ DBLP:journals/corr/abs-1211-6581,Borchani2015ASO}, the proposed model processes all targets simultaneously without splitting them into different single-output models. Moreover, we contribute to the overall goal of transparent and safer machine learning models in the following ways.
\begin{itemize}
\item We propose a new regression model at the intersection of classical methods and deep PNN. It performs comparably to the state-of-the-art models and outperforms them on specific tasks.
\item The model is theoretically grounded and closely relates to the theory of ODEs, which opens up the possibility of model interpretation and analysis.
\item We provide an extensive evaluation of the model against both classical and state-of-the-art methods on a set of 33 regression UCI open access datasets, the Feynman symbolic regression benchmark with 120
datasets, and the Friedman-1 dataset.
\item We provide detailed analyses to understand model’s performance: ablation study, the impact of data properties with different numbers of unimportant features, noise levels, and number of samples.
\end{itemize}

\section{Method}
To describe the proposed regression model for the problem \eqref{reg}, let us first define a Taylor map as the transformation $\MM : \ZZ_{t}\rightarrow \ZZ_{t+1}$ in the form of
\begin{equation}
	\label{tmap}
	\MM: \ZZ_{t+1} = W_0 + W_1\,\ZZ_{t}+\ldots+W_k\,\ZZ_{t}^{[k]},
\end{equation}
where $\ZZ_{t}, \ZZ_{t+1} \in \mathcal R^{n+m}$, matrices $W_0, W_1,\ldots,W_k$ are trainable weights, and $\ZZ^{[k]}$ means $k$-th Kronecker power of vector $\ZZ$. The transformation \eqref{tmap} can be referred to Taylor maps and models \cite{ref_tm1}, exponential machines \cite{ref201}, tensor decomposition \cite{ref_tm2}, and others. In fact, map $\MM$ is just a multivariate polynomial regression of $k$-th order.

\subsection{Proposed model}
Let's now extend feature vector $\XX = \{\x_1,\ldots,\x_n\}$ in the regression problem \eqref{reg} with additional $m$ dimensions filled with zeros and consistently apply the map \eqref{tmap} $p$ times:
\begin{equation}
\label{tmreg}
\begin{split}
    \ZZ_0 &= \{\x_1,\ldots,\x_n,0,\ldots,0\},\\
    \ZZ_1 &= W_0 + W_1\,\ZZ_0+\ldots+W_k\,\ZZ_{0}^{[k]},\\
    \ldots\\
    \ZZ_{p} &= W_0 + W_1\,\ZZ_{p-1}+\ldots+W_k\,\ZZ_{p-1}^{[k]}.
\end{split}
\end{equation}
This procedure implements Taylor mapping that propagates initial vector $\ZZ_0$ along new discrete dimension $t=\overline{1,p}$ with hidden states
$
\ZZ_t =
\{x_{1,t},\ldots,x_{n,t},\,y_{1,t}, \ldots,y_{m,t}\}
$.

For the final state $\ZZ_p$ this yields a polynomial regression of order $k^p$ with respect to the components of $\ZZ_0$ which is, however, factorized by a Taylor map $\MM$ of order $k$:
\begin{equation*}
\ZZ_{p} = \MM \circ \ldots \circ \MM \circ \ZZ_{0} = V_0 + V_1\,\ZZ_0+\ldots+V_{k^p}\,\ZZ_{0}^{[k^p]},
\end{equation*}
where weight matrices $V_q = V_q(W_0, W_1,\ldots,W_k)$ for $q=\overline{0,k^p}$. 
For defining a loss function, one should consider only the last $m$ components of the vector $\ZZ_p = \{x_{1,p},\ldots,x_{n,p},\,y_{1,p}, \ldots,y_{m,p}\}$ as the predictions:
\begin{equation}
\label{loss}
Loss = \rho(\{\y_1, \ldots,\y_m\}, \{y_{1,p}, \ldots,y_{m,p}\}),
\end{equation}
where $\rho$ is an error between true targets $\{\y_1, \ldots,\y_m\}$ and predicted values $\{y_{1,p}, \ldots,y_{m,p}\}$.

\begin{figure}
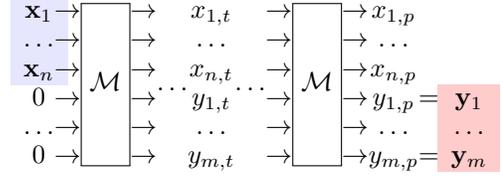

\vskip 0.2in
\begin{center}
\begin{equation*}
    \begin{matrix}
    \colb \x_1\\
    \colb \ldots\\
    \colb \x_n\\
    0\\
    \ldots\\
    0\\
    \end{matrix}
     \begin{matrix}
    \rightarrow\\
    \rightarrow\\
    \rightarrow\\
    \rightarrow\\
    \rightarrow\\
    \rightarrow\\
    \end{matrix}
    \boxed{
    \begin{matrix}
    \\
    \\
    \MM\\
    \\
    \\
    \end{matrix}
    }
    \begin{matrix}
    \rightarrow\\
    \rightarrow\\
    \rightarrow\\
    \rightarrow\\
    \rightarrow\\
    \rightarrow\\
    \end{matrix}
    \begin{matrix}
    \ldots
    \end{matrix}
    \begin{matrix}
    x_{1,t}\\
    \ldots\\
    x_{n,t}\\
    y_{1,t}\\
    \ldots\\
    y_{m,t}\\
    \end{matrix}
    \begin{matrix}
    \ldots
    \end{matrix}
    \begin{matrix}
    \rightarrow\\
    \rightarrow\\
    \rightarrow\\
    \rightarrow\\
    \rightarrow\\
    \rightarrow
    \end{matrix}
    \boxed{
    \begin{matrix}
    \\
    \\
    \MM\\
    \\
    \\
    \end{matrix}
    }
    \begin{matrix}
    \rightarrow\\
    \rightarrow\\
    \rightarrow\\
    \rightarrow\\
    \rightarrow\\
    \rightarrow\\
    \end{matrix}
    \begin{matrix}
    x_{1,p}\\
    \ldots\\
    x_{n,p}\\
    y_{1,p}\\
    \ldots\\
    y_{m,p}\\
    \end{matrix}\begin{matrix}
    \\
    \\
    \\
    =\;\,\,\\
    \\
    =\;\,\,\\
    \end{matrix}
    \begin{matrix}
    \\
    \\
    \\
    \colr \y_1\\
    \colr \ldots\\
    \colr \y_m\\
    \end{matrix}
\end{equation*}
\caption{High-order regression  $\{\x_1, \x_2, \ldots, \x_n\}\;\rightarrow\;\{\y_1, \y_2, \ldots, \y_m\}$ factorized by Taylor map $\mathcal M$ and implemented as PNN with $p$ layers of shared weights.}
\label{fig1_tmpnn}
\end{center}
\vskip -0.2in
\end{figure}

Fig.~\ref{fig1_tmpnn} presents this algorithm as a Taylor map-based PNN (TMPNN) with layers of shared weights that propagate the extended vector of features  $\ZZ_0$. Namely, after the first layer, all variables in the feature vector are modified, and the variables on the $m$ introduced dimensions become non-zero in general. The mapping \eqref{tmap} $\ZZ_t \rightarrow \ZZ_{t+1}$ continues until the $p$-th layer, where the last $m$ variables are used for the output.

If the number of layers $p=1$, the proposed architecture is equivalent to regular polynomial regression \eqref{polreg} of order $k$. If the order of the Taylor map $k=1$, then the proposed model is equivalent to linear regression. The proposed architecture defines a high-order polynomial regression of order $k^p$ for other cases. Since the weights of $\MM$ are shared, the number of free parameters in the model equals to 
$(n+m)\sum_{l=0}^{k}C_{n+m-1+l,n+m-1},$
where a combination $C_{a+b,a}$ defines the number of the monomials of degree $b$ with $a+1$ variables. This model can result in extremely high-order polynomials with significantly fewer free parameters than regular polynomial regression. 


The architectures of polynomial transformations
with shared weights have been previously used in various research topics but were not formulated for general-purpose regression. For example, \cite{DragtKick} used Taylor maps to factorize dynamics in Hamiltonian systems. \cite{TMA} applied the same concept for data-driven control of X-ray sources. \cite{pinet} described a similar computational graph with polynomial shared layers for applications in computer vision. The current paper considers the algorithm \eqref{tmreg} as a general-purpose regression model for multi-target problems.

\begin{figure*}[t]
\centering
\includegraphics[width=1\textwidth]{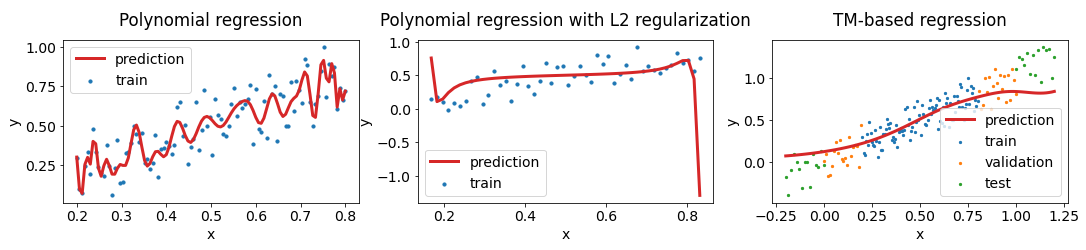} 
\caption{Comparison of regular 41st order polynomial regression (left plot), regularized one (central plot), and proposed  regression of 125th order (right plots) for the fitting of noisy linear data. Each of the models has 41 weight coefficients.}
\label{linearmodelcomparison}
\end{figure*}

\subsection{Model hyperparameters}
The proposed model has two fundamental hyperparameters: the order $k$ of the nonlinearities in the map \eqref{tmap} and the number $p$ of iterations in \eqref{tmreg}. These parameters define the order $k^p$ of the polynomials that represent the relationship between the input variables and targets.

We speculate that in addition to defining the order $k^p$ of the polynomial, the number of steps $p$ also serves as a regularization. More iterations in \eqref{tmreg} result in larger derivatives for the high-order terms in the loss. While we can claim empirical evidence for this behavior, the theoretical properties should be examined in detail.
Moreover, one can introduce classical regularization, such as L1, L2, and dropout, and incorporate their respective hyperparameters.

Finally, the initial values $\{y_{1,0}, \ldots,y_{m,0}\}$ in $\ZZ_0$ can be considered as free parameters. Instead of extending the input vector $\XX$ with zeros, it is possible to consider any other values, resulting in a new set of $m$ hyperparameters.

\subsection{Implementation}
A naive implementation of the proposed regression model can be done in Python using a simple \texttt{\textbf{for}} loop. An example of the second-order mapping \eqref{tmap} is presented in Listing \ref{lst:naiveimpl}.

More advanced implementation using Keras and TensorFlow is provided in the supplementary code. It includes a layer that implements a Taylor map \eqref{tmap} of arbitrary order of nonlinearities and the construction of the computational graph \eqref{tmreg}. We use the Adamax optimizer \cite{DBLP:journals/corr/KingmaB14} with default parameters based on our experiments. The paper does not cover the choice of the optimal optimization method for the proposed model.

\begin{listing}[h]%
\caption{Proposed regression for $k=2$ and $p=10$}%
\label{lst:naiveimpl}%
\begin{lstlisting}[language=Python]
def predict(X, W0, W1, W2, num_targets):
    num = X.shape[0] # number of samples
    Y0 = np.zeros((num, num_targets))
    # extend X with zeros:
    Z = np.hstack((X, Y0))
    for _ in range(10): # p=10
        # 2nd kronecker power:
        Z2 = (Z[:,:,None]*Z[:,None,:])
        Z2 = Z2.reshape(num, -1)
        Z = W0 + np.dot(Z, W1) + np.dot(Z2, W2)
    # last num_targets values:
    return Z[:, -num_targets:]
\end{lstlisting}
\end{listing}

\subsection{Example}
To demonstrate the main difference of \eqref{tmreg} in comparison to the regular polynomial regression \eqref{polreg}, the linear model
$
\y = \x + \epsilon
$
with a single variable $\x$, a single target $\y$, and random noise $\epsilon \sim U(-0.25, 0.25)$ is utilized. 

For the regular polynomial regression \eqref{polreg}, we use a polynomial of 41-st order just for example
\begin{equation}
\label{linearpoly}
\hat{\y}_{pred} = c_0 + c_1\x + \ldots + c_{41}\x^{41},
\end{equation}
and try using both non-regularized and L2-regularized mean squared error (MSE).

For the proposed model \eqref{tmreg},
we use fifth order map \eqref{tmap} and three steps in \eqref{tmreg}. This results in 125-th order polynomial $\y_{pred} = y_3 $ with $x_0 = \x$ and $y_0 = 0$ incorporating the same number of free parameters:

\begin{equation}
\label{lineartmpnn}
\begin{split}
&\begin{pmatrix}
    x_{t+1}\\
    y_{t+1}\\
\end{pmatrix}
=\begin{pmatrix}
    w_0\\
    w_1\\
\end{pmatrix}
+
\begin{pmatrix}
    w_2 & w_3\\
    w_4 & w_5\\
\end{pmatrix}
\begin{pmatrix}
    x_t\\
    y_t\\
\end{pmatrix}\\
&+\begin{pmatrix}
    w_6 & w_7 & w_8\\
    w_{9} & w_{10} & w_{11}\\
\end{pmatrix}
\begin{pmatrix}
    x_t^2\\
    x_ty_t\\
    y_t^2\\
\end{pmatrix}+\ldots\\
&+\begin{pmatrix}
    w_{30} & w_{31} & \ldots & w_{35}\\
    w_{36} & w_{37} & \ldots & w_{41}\\
\end{pmatrix}
\begin{pmatrix}
    x_t^5\\
    x_t^4y_t\\
    \ldots\\
    y_t^5\\
\end{pmatrix}.
\end{split}
\end{equation}

As expected, polynomial regression \eqref{linearpoly} leads to overfitting, resulting in a curve with many kinks attempting to predict the noise.
L2 regularization shrinks all coefficients $c_i$ to zero but still does not help with poor extrapolation (see Fig.~\ref{linearmodelcomparison}, left and central plots).

\begin{figure}[h]
\centering
\includegraphics[width=0.85\columnwidth]{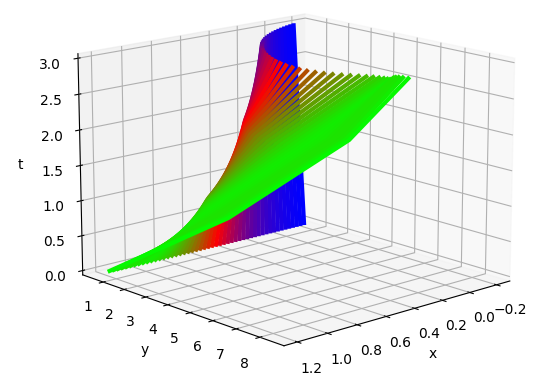}
\caption{Surface fitted in the space $(x_t, y_t, t)$ with the proposed high-order regression \eqref{tmreg} for linear data.}
\label{lineartmpnn3d}
\end{figure}

The proposed model, instead of fitting one-dimensional curves, attempts to fit a surface in the three-dimensional space $(x_t, y_t, t)$. Starting with curve ($\x, 0, 0)$, the model propagates it to the $(x_3, \y_{pred}, 3)$. The visualization of this approach is provided in  Fig.~\ref{lineartmpnn3d}. The fitted model \eqref{lineartmpnn} without any extra regularization has coefficients $w_i$ ranging from -0.075 to 0.11. As a result, the final high-order polynomial is smoother, providing a better extrapolation for a wider region of unseen data (see Fig.~\ref{linearmodelcomparison}, right plot).

\section{Iterpretability and Approximation}
\label{sec:interp}
In this section, we demonstrate that the TMPNN can be considered a numerical approximation of the general solution for a system of ODEs that approximates problem \eqref{reg}. Note that although the TMPNN is theoretically grounded with ODEs, the model, as a PNN with shared weights, eliminates the need for numerical solvers for ODEs during training. 


\subsection{Interpretation}
\label{sec:interpretation}
The Taylor map \eqref{tmap}  can be written in the form:
\begin{equation}
\label{tmap2}
\begin{split}
\ZZ_{t+1} = \ZZ_t &+ \Delta\tau\left(pW_0 + p(W_1-I)\ZZ_t\right.\\
&+\left.pW_2\,\ZZ_{t}^{[2]}
+\ldots+pW_k\,\ZZ_{t}^{[k]}\right),
\end{split}
\end{equation}
with $\Delta\tau = 1/p$ and $I$ for identity matrix. Comparing \eqref{tmap2} with the simplest Euler method for solving ODEs with a right-hand side $F$:
\begin{equation}
\begin{split}
\frac{d}{d\tau}\ZZ(\tau) &= F\left(\ZZ(\tau)\right),\\
\ZZ(\tau + \Delta\tau) &= \ZZ(\tau) + \Delta\tau F\left(\ZZ(\tau)\right),
\end{split}
\end{equation}
one can conclude that Taylor map \eqref{tmap} represents the difference equations for a discrete state $\ZZ_t$ for the evolution of the state vector $\ZZ(\tau) = (x_1(\tau), \ldots, x_n(\tau), y_1(\tau), \ldots, y_m(\tau))$ in continuous time $\tau \in [0, 1]$ with time step $\Delta \tau = 1/p$:
\begin{equation}
\label{odei}
    \frac{d}{d\tau}
    \begin{pmatrix}
    x_1(\tau)\\
    \ldots\\
    x_n(\tau)\\
    y_1(\tau)\\
    \ldots\\
    y_m(\tau)\\
    \end{pmatrix}
    = A_0 + A_1
    \begin{pmatrix}
    x_1(\tau)\\
    \ldots\\
    x_n(\tau)\\
    y_1(\tau)\\
    \ldots\\
    y_m(\tau)\\
    \end{pmatrix}
    + A_2
    \begin{pmatrix}
    x_1(\tau)\\
    \ldots\\
    x_n(\tau)\\
    y_1(\tau)\\
    \ldots\\
    y_m(\tau)\\
    \end{pmatrix}^{[2]}+\ldots,
\end{equation}
where $A_q = pW_q$ for $q\neq1$ and $A_1 = p(W_1 - I)$.

Furthermore, the proposed model \eqref{tmreg} corresponds to the boundary value problem \eqref{odei} with the initial conditions
\begin{equation}
\label{initcond}
\begin{split}
&x_1(0) = \x_1, x_2(0)=\x_2, \ldots, x_n(0) = \x_n,\\
&y_1(0) = 0, \ldots, y_m(0) = 0
\end{split}
\end{equation}
and the boundary conditions
\begin{equation}
    \label{bouncond}
    y_1(\tau=1) = \y_1, \ldots, y_m(\tau=1) = \y_m.
\end{equation}

Otherwise stated, the regression model  \eqref{tmreg} with trainable weights $W_q$ is equivalent to the boundary value problem \eqref{initcond} and \eqref{bouncond} for the system of ODEs \eqref{odei} with trainable weights $A_q$. The particular solution $\ZZ(\tau)$ of the system \eqref{odei} for the initial conditions \eqref{initcond} at time $\tau=t/p$ corresponds to the discrete state $\ZZ_t$ after the $t$-th step in the \eqref{tmreg}. 

The connection between the proposed model \eqref{tmreg} and ODEs highlights that both features and targets, as well as all targets together, are coupled. It provides a natural way to construct a multi-target regression without building multiple single-output models. Moreover, since the trainable weights $W_q$ correspond to $A_q$ in the system \eqref{odei}, this approach allows us to reconstruct the system of ODEs that approximates the training dataset without the need to solve differential equations numerically. Only the weights in \eqref{tmreg} are tuned with the data during training.


\begin{table*}[h]
    \centering
    \begin{tabular}{ccccccccc} \hline
         & Sarcos & CO$^2$Emission & California &  Airfoil & Concrete & NOxEmission & Pendulum\\ \hline
        CatBoost &1.71$\pm$0.06& \textbf{ 1.02$\pm$0.13}  & \textbf{ 0.19$\pm$0.01} & \textbf{1.34$\pm$0.30} & \textbf{15.34$\pm$7.66} & 14.73$\pm$1.12 & 4.55$\pm$2.58\\ 
        DNNR &\textbf{0.80$\pm$0.05 }& 1.28$\pm$0.13 & 0.24$\pm$0.02 & 4.56$\pm$0.86 & 35.98$\pm$9.01 & 18.39$\pm$1.78 & 2.83$\pm$1.64 \\ 
        TabNet & 1.56$\pm$0.25 & 1.12$\pm$0.21 & 0.39$\pm$0.03 & \textbf{ 1.36$\pm$0.39} & 18.50$\pm$6.27 & \textbf{11.17$\pm$1.25} & 2.40$\pm$1.10\\
        XGBoost & 2.12$\pm$0.06 & 1.14$\pm$0.14 & 0.21$\pm$0.01 & 1.62$\pm$0.33 & \textbf{16.90$\pm$7.92} & 16.21$\pm$0.89 & 5.09$\pm$2.68\\
        MLP & 1.17$\pm$0.12 & 1.08$\pm$0.21 & 0.30$\pm$0.02 & 6.18$\pm$1.57 & 22.72$\pm$7.57 & 14.78$\pm$1.57 & \textbf{1.80$\pm$0.98} \\
        KNN & 2.47$\pm$0.09 & 1.10$\pm$0.18 & 0.40$\pm$0.01 & 8.26$\pm$1.06 & 70.73$\pm$13.63 & 18.54$\pm$1.10 & 4.03$\pm$2.03 \\
        \hline
        TMPNN & 1.40$\pm$0.05 & 1.27$\pm$0.26 & 0.38$\pm$0.01 & 3.15$\pm$1.06 & 26.46$\pm$4.47 & 17.30$\pm$4.90 & 2.45$\pm$1.10 \\\hline
    \end{tabular}
    \caption{ The MSE on various UCI datasets averaged over random 10-fold cross-validation. The standard deviations are given after the $\pm$ signs.}
    \label{randomCV}
\end{table*}

\subsection{Approximation capabilities}

\textbf{Technical assumptions:} The feature space $\mathcal{X} \subset \mathcal{R}^n$ forms a compact Hausdorff space (separable metric space). The feature distribution $\mathbb{P}_\XX$ is a Borel probability measure.

\noindent\textbf{Besicovitch assumption:} The regression function $\eta:\mathcal{X}\rightarrow\mathcal{R}^m$ satisfies $lim_{r\to+0}\mathbb{E}\left[\YY|\XX\in B_{\x,r}\right]=\eta(\x)$ for $\x$ almost everywhere w.r.t. $\mathbb{P}_\XX$, with $B_{\x,r}$ standing for a ball with radius $r$ around $\x$.

\noindent\textbf{Lemma:} For any function $\eta\in \mathcal{L}_p$ (Lebesgue spaces) and tolerance $\forall\varepsilon>0$ a continuously differentiable function $f\in C^1$ exists such that a norm in $\mathcal{L}_p$ $||\eta(\x)-f(\x)||_p\leq\varepsilon$ under technical assumptions. See prof in \cite{lpdense}.


\noindent\textbf{Theorem:}
\textit{Under the given assumptions and lemma, for any function $\eta\in\mathcal{L}_p$ and tolerance $\forall\varepsilon>0$ parameters $k$ and $p$ in the model \eqref{tmreg} exist such that $||y(p) - f(\x)||_p\leq\varepsilon$}.


To prove this theorem, without loss of generality, we consider the example of a scalar function $\y = \eta(\x)$ of one variable. Let's approximate it with a given tolerance $\varepsilon/2$ with continuously differentiable function $f(\x)$. By introducing two variables $x(\tau)$ and $y(\tau)$, one can derive a system of ODEs that implements mapping $y(0)=0 \rightarrow y(1) = f(\x)$. For this one can suppose $y = \xi(\tau, x)f(x)$ with conditions $ \xi(0, x)  \equiv 0, \xi(1, x)  \equiv 1$, and write
the differential of $\xi$:
$$
\frac{\partial \xi}{\partial \tau}d\tau + \frac{\partial \xi}{\partial \x}d\x = (fdy - y \frac{\partial f}{\partial \x}dx)/f^2.
$$
Using this relation, one can write a system
\begin{equation}
\label{odei2}
\begin{split}
f(x)\frac{dy}{d\tau} = \frac{\partial \xi}{\partial \tau}f^2(x) + \Theta(x, y, \tau),\\
 \left(y \frac{\partial f}{\partial \x} + f^2(x)\frac{\partial \xi}{\partial \x} \right) \frac{dx} {d\tau} = \Theta(x, y, \tau),
\end{split}
\end{equation}
where $\Theta(x, y, \tau)$ is chosen arbitrary.
Equations \eqref{odei2} can then be written as the system of ODEs by choosing $\Theta(x, y, \tau) = \nu(x,y,\tau) \left(f(x) y \frac{\partial f}{\partial \x} + f^2(x)\frac{\partial \xi}{\partial \x}\right)$:
\begin{equation}
\label{approxode}
    \begin{split}
        \frac{dx}{d\tau} &= \nu f(x),\\
        \frac{dy}{d\tau} &= \frac{\partial \xi}{\partial \tau}f(x) + \nu f^2(x)\frac{\partial \xi}{\partial x} + \nu y \frac{\partial f(x)}{\partial x},
    \end{split}
\end{equation}
which together with the initial conditions 
$ x(0) = \x, y(0) = 0$
leads to the desired solution $y(1) = f(\x)$. Note that $\nu$ and $\xi$ can be chosen such that the system \eqref{approxode} is autonomous.

The system of ODEs \eqref{approxode} represents one of many possible forms that the proposed model \eqref{tmreg} implicitly learns in its discrete step-by-step representation.
Moreover, in the case of  $\nu=0, \xi=\tau$, the system \eqref{approxode} corresponds to the regular polynomial regression \eqref{polreg}. Indeed, in this case
\begin{equation}
\label{approxode2}
    \begin{split}
        \frac{dx}{d\tau} = 0,\;\;\;\;
        \frac{dy}{d\tau} = f(x).
    \end{split}
\end{equation}
If the function $f(\x)$ in \eqref{approxode2} is a polynomial of $k$-th order, the Taylor map
$
y_{t+1} = y_{t} + f(x(0))/p
$
with the initial condition $x(0) = \x$, $ y_0 = y(0) = 0$ exactly produces regular polynomial regression $\y = y_p = f(\x)$.
For the non-polynomial functions $f(\x)$, the systems \eqref{approxode} still corresponds to the output $\y = y(1) = f(\x)$, but the Taylor mapping \eqref{tmreg} approximates this solution under technical assumptions with an arbitrate accuracy $\varepsilon/2$ depending on the order $k$ and the number of steps $p$ due to the Stone-Weierstrass theorem \cite{Stone, Weierstrass}. Thus, the Taylor mapping also approximates a given regression function $\eta$ with $\varepsilon$ accuracy. The convergence and accuracy of \eqref{tmreg} with respect to parameters $k$ and $p$ can be estimated theoretically
\cite{MFCONV, UIben}. This proves the theorem.


\subsection{Increasing the order of polynomial regression}

To generate a higher order polynomial regression, one can increase order $k$ or add steps $p$ in \eqref{tmreg}. Rising the order increases the complexity of the model while increasing steps preserve the same number of trainable parameters but make TMPNN deeper that can affect optimization performance without proper weights initialization. 

Section \ref{sec:interpretation} provides an approach for the weights initialization. Given the model \eqref{tmreg} with parameters $k$, $p$, and trained weights $W_i$, one can write the system \eqref{odei} and generate a new model \eqref{tmreg}  by integrating \eqref{odei} with the increased number of steps $\bar p > p$. This results in the new weights
\begin{equation}
\label{increasep}
\begin{split}
\bar W_q &= pW_q/\bar p,\;\;  q\neq 1,\\
\bar W_1 &= pW_q/\bar p + (\bar p - p)I/\bar p
\end{split}
\end{equation}
for the TMPNN with more steps and, as a result, higher polynomial regression of order $k^{\bar p}$. This approach transforms lower-order model $k,p$ to a higher-order $k,\bar p$, providing a proper weight initialization. We only introduce this possibility without an experiment study of the approach in the paper.

\section{Experiments}

The experiments setup, hyperparameters for the considered models, and code can be found in supplementary materials for reproducibility. 

\subsection{Interpolation}

\subsubsection{Feynman Benchmark}
Feynman Symbolic Regression Database consists of 120 datasets sampled from classical and quantum physics equations \cite{Feynman}. These equations are continuous differentiable functions. We increased the difficulty of the datasets by adding Gaussian noise with three standard deviations (0, 0.001, 0.01). The evaluation was executed
with ten different random splits (75\% for training and 25\% for testing and metric reporting) for each dataset and noise level.

\begin{figure}[h]
\centering
\includegraphics[width=0.95\columnwidth]{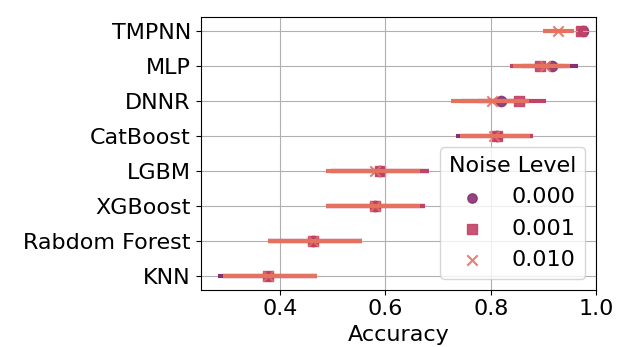}
\caption{Accuracy shows the percentage of solutions with R2 $>$ 0.999 on the Feynman Regression dataset under three noise levels. The bars denote 95\% confidence intervals.}
\label{feynman}
\end{figure}

We define the accuracy as the percentage of datasets solved with a coefficient of determination R2 $>$ 0.999 and report it in Fig.~\ref{feynman}. The accuracy of the classical and state-of-the-art models was taken from \cite{DNNR}, where optimal hyperparameters search was performed for the same datasets. We do not use optimal hyperparameter search for the proposed model and suppose $k=3, p=5$ as default. The proposed model is the best-performing one with the smallest confidence intervals independently of the noise level.

\subsubsection{UCI datasets}
Table~\ref{randomCV}  demonstrates the performance of the proposed model (TMPNN) for several UCI\footnote{ https://github.com/treforevans/uci\_datasets to access datasets} 
and Sarcos open access datasets. For evaluation, we used several state-of-the-art and classical models. CatBoost is the best-performing method. TMPNN demonstrates comparable results against other models serving as the second and third-best model for the Pendulum and Sarcos datasets.

For the remaining 32 UCI regression datasets with less than 50 features, we compare TMPNN with $k=2,p=7$, CatBoost, and DNNR with default parameters. For the evaluation of each model, we use only datasets with $R2>0.5$. CatBoost and TMPNN provide the same average $R2 = 0.88 \pm 0.14$ on 26 datasets. DNNR results in average $R2 = 0.87 \pm 0.12$ but fits only 18 datasets.
\begin{figure*}[t]
\centering
\includegraphics[width=1\textwidth]{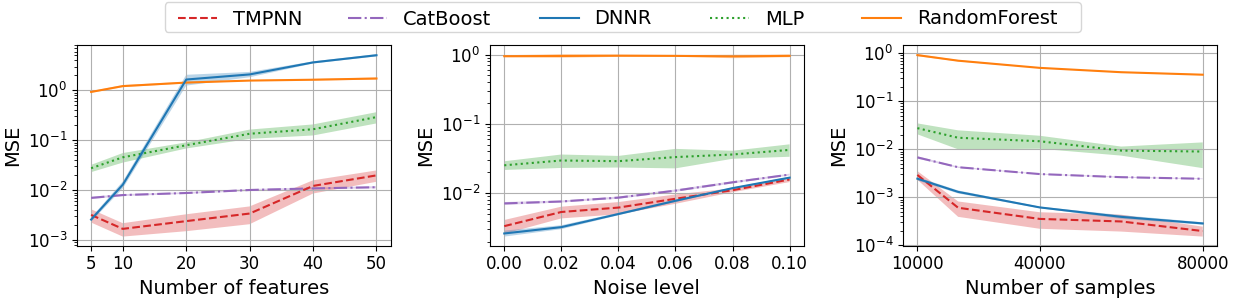} 
\caption{The effect of different sampling conditions on model performance for the Friedman-1 dataset.}
\label{Feynman}
\end{figure*}

\subsection{Ablation study}
\label{sec: ablation}
In this section, we discuss TMPNN performance under various design alternatives, such as initialization of weights, number of layers $p$ and nonlinear transformations $k$.  We base this analysis on the Airfoil and 10000 samples from Friedman-1 datasets \cite{Friedman}. 

Initializations differed from identity mapping \eqref{tmap} lead to worse results for both datasets. Increase of layers starting at $p=5$ results in a decrease in performance. We suppose that this affects the optimizer because  we do not use relation \eqref{increasep} for weights initialization.

\begin{table}[ht]
\centering
\begin{tabular}{cccc}
\hline
&&Airfoil&Friedman-1\\
\hline
&k=2, p=5 & $0.87\pm0.01$& $0.99\pm0.00$\\
$W_i=0$&k=3, p=5 & $0.92\pm0.01$& $0.99\pm0.00$\\
$i\neq1$&k=3,p=10 & $0.89\pm0.05$&$0.99\pm0.00$ \\
$W_1=I$&k=3,p=20 & $0.87\pm0.11$& $0.99\pm0.00$\\
&k=4, p=5 & $0.80\pm0.12$&$0.99\pm0.00$ \\
\hline
&k=2, p=5 & $0.66\pm0.02$& $0.99\pm0.00$\\
$W_i=0$&k=3, p=5 & $-1.07\pm2.44$&$0.99\pm0.00$ \\
&k=4, p=5 & $-3.86\pm4.29$& --\\
\hline
$W_i=\epsilon_i,$&k=2, p=5 & $0.86\pm0.02$&$0.99\pm0.00$\\
$i\neq1$&k=3, p=5 & $0.91\pm0.1$&$0.99\pm0.00$ \\
$W_1=I+\epsilon_1$&k=4, p=5 & $0.74\pm0.24$& -- \\
\hline
\end{tabular}
\caption{R2 score for variations of TMPNN. $\epsilon_j\subset \mathcal N (0, 0.0001)$. Sign "--" stands for the failure to converge.}
\label{tab:competition}
\end{table}

\subsection{Effect of noise, number of samples and features}

This section investigates how the noise, the number of samples, and presence of unimportant features affect the model’s performance. Since such an analysis requires a controlled environment, we used the classical Friedman-1 dataset that allows varying sampling conditions. Friedman-1 datasets are generated for
five uniformly distributed in $[0, 1]$ features, the noise sampled from a standard normal distribution. Additional unimportant features are added
by sampling from the continuous uniform distribution $U(0, 1)$.

Besides TMPNN, we also evaluated CatBoost, DNNR, MLP, and Random Forest. We use the same train-test splitting as for the Feynman benchmark. The hyperparameters for each method are fitted on the default dataset (10000 samples, five features, without noise) and are fixed for the remaining analysis. Fig.~\ref{Feynman} reports the effects of each condition. TMPNN performs the best for the low number of added unimportant features but is beaten by CatBoost slightly for higher numbers. TMPNN is the only model that demonstrates error reduction when adding a small number of unimportant features. This behavior refers to the extra latent dimensions and is discussed briefly in Section~\ref{sec:further}. For the noise level, CatBoost, DNNR, and TMPNN  perform similarly. For the number of samples, TMPNN is the best model outperforming even DNNR, which is based on nearest neighbors.

\subsection{Extrapolation}

\subsubsection{UCI Yacht Hydrodynamics dataset}

The UCI Yacht Hydrodynamics dataset consists of 308 experiments connecting six features (lcg, cp, l/d, b/d, l/b, fn) with a single target (rr) \cite{yacht}. To compare the extrapolation ability of different models, we utilize three models using cross-validation for optimal hyperparameters choice.


We use several extrapolation train-test splits when samples exceeding 75\% quantile by each feature and target are chosen for test sets. Table~\ref{tableUCI} presents R2 scores for each of the testing datasets. The first row refers to the rule for test dataset generation and its size in percentage. CatBoost, DNNR and MLP are unsuitable for out-of-distribution prediction depending on feature fn and target rr. In these cases, the test set contains completely different values for the target. Regular polynomial regression and TabNet perform better but still unstable. TMPNN provides the best scores for each split.

\begin{table}[ht]
\centering
\begin{tabular}{ccccc}
\hline
&cp$>$0.5& b/d$>$4.1 & fn$>$0.4 & rr$>$12\\
&(22\%)& (22\%) & (21\%) & (25\%)\\
\hline
MLP&  0.94 & 0.88 &  -0.53 & -0.49 \\
CatBoost& 0.98 & 0.97 & -3.27 &-2.40\\
DNNR& \textbf{0.99} & \textbf{0.99} & -0.46 & -0.56 \\
Polyn. Regr.& 0.63 & -0.61 & 0.89 & 0.83 \\
TabNet& \textbf{0.99} & \textbf{0.99} &  0.71 & -0.01\\
\hline
TMPNN& \textbf{0.99} & \textbf{0.99} & \textbf{0.92} & \textbf{0.90} \\
\hline
\end{tabular}
\caption{R2 scores on various extrapolation tests for different models on UCI Yachr Hydrodynamics dataset.}
\label{tableUCI}
\end{table}

\section{Further development}
\label{sec:further}


Theoretical results from fields of PNNs and ODEs can be utilized to estimate the error of the proposed model and explore the convergence of the training process. Further theoretical investigation is required to understand the relationship between the sufficient number of training steps, the hyperparameters, and the presence of noise in the data.

The extrapolation ability of the proposed model can be explored by drawing upon the theory of ODEs. As the model implicitly learns the general solution of a system of ODEs, the out-of-distribution inference can be viewed as new initial conditions for already learned general solution.

Due to the Picard-Lindelöf theorem for ODEs, the proposed model can be utilized for classification problems constraining the trajectories $\ZZ_t$ to converge to the target states $\ZZ_c\in \mathcal R^{n+m}$ for each class $c$. While the applicability of PNNs for classification has been discussed in \cite{DBLP:journals/corr/abs-2104-07916}, it should be extended to the proposed model.

The recurrent form of Taylor mapping also makes it possible to apply in multivariate time-series analysis. The internal states $\ZZ_t$ can be represented as time-series counts. Moreover, the model can be used for sparse time-series forecasting. Since the model can be considered as integration of a system of ODEs \eqref{odei} over time, the required time period $\bar \tau$ might be changed fluently with $\bar W_q = W_q\bar \tau/\tau$ for $q\neq1$ and $\bar W_1 = (W_1 - I)\bar \tau/\tau + I$ with $I$ for identity matrix and $\tau=1$ for normalization.



Regarding feature engineering, there are two possibilities for improving the model. Firstly, one can extend the state vector  $\ZZ$ by appending
$l$ latent units resulting in $\ZZ \rightarrow \ZZ_E \in \mathcal R^{n+m+l}$. This increases the expressivity of the model and has been successfully applied in regression tasks in prior works \cite{ref9, odepkr}. Secondly, instead of initializing the state $\ZZ_0$ with the last $m$ zeros, one can use learnable parameters, custom functions, or predictions from other models.

While TMPNN provides a way to build a polynomial regression of order $k^p$, the approach \eqref{increasep} allows increasing the order of polynomial even more. The iterative weights initialization helps to increase the number of layers $p$, resulting in a higher polynomial. This approach requires further investigation with datasets.





The possible limitations of the TMPNN need to be considered in more detail. These include the influence on data normalization,
presence of categorical and ordinal features, and the differentiability of the functional relationship between targets and features. Investigating these aspects is crucial for effectively applying the model to diverse datasets.

\section{Conclusion}



As mentioned above, certain theoretical investigations regarding the proposed model were not addressed within this paper's scope and remain open to further research.
The main goal of the current paper is presenting a novel approach for constructing extremely high-order polynomials to solve regression problems and empirically demonstrate its validity.





There are two main contributions in the paper. Firstly, we introduced a high-order polynomial model designed to accommodate multi-target regression. 
The key for understanding of TMPNN is replacing regular 'curve fitting' $\x \rightarrow \y$  with 'surface fitting' in the extended space $ (x,y,t): (\x, 0, 0) \rightarrow (x(1), \y, 1)$.
Secondly, we presented the interpretation and approximation capabilities of the model by establishing its connection with ODEs.

With the example of 155 datasets and different sampling conditions such as noise level, number of samples, and presence of unimportant features, we demonstrated that the proposed model is well-suited for regression tasks and outperforms state-of-the-art methods on specific problems. 
Moreover, connection of the TMPNN with ODEs makes it naturally applicable to various domains such as physical, chemical, or biological systems. At the same time, TMPNN does not require numerical ODEs solvers during training. 



\section{Code}
The implementation of the model in Python and Keras/TensorFlow is available in \url{https://github.com/andiva/tmpnn}.

\bibliography{aaai24}

\end{document}